# Exploring Large Language Models for Climate Forecasting

Yang Wang and Hassan A. Karimi

*Abstract*— With the increasing impacts of climate change, there is a growing demand for accessible tools that can provide reliable future climate information to support planning, finance, and other decision-making applications. Large language models (LLMs), such as GPT-4o, present a promising approach to bridging the gap between complex climate data and the general public, offering a way for non-specialist users to obtain essential climate insights through natural language interaction. However, an essential challenge remains underexplored: evaluating the ability of LLMs to provide accurate and reliable future climate predictions, which is crucial for applications that rely on anticipating climate trends. In this study, we investigate the capability of GPT-4o in predicting rainfall at short-term (15-day) and long-term (12-month) scales. We designed a series of experiments to assess GPT's performance under different conditions, including scenarios with and without expert data inputs. Our results indicate that GPT, when operating independently, tends to generate conservative forecasts, often reverting to historical averages in the absence of clear trend signals. This study highlights both the potential and challenges of applying LLMs for future climate predictions, providing insights into their integration with climate-related applications and suggesting directions for enhancing their predictive capabilities in the field.

## I. INTRODUCTION

As the impacts of climate change intensify, obtaining accurate information about future climate trends has become increasingly important. Fields such as energy planning, urban development, and weather derivatives increasingly rely on precise climate forecasts to make critical decisions [1]. However, accessing, analyzing, and interpreting climate data typically requires interdisciplinary knowledge in areas such as climatology, geography, statistics, and computer science, making it challenging for the general public to utilize this information effectively. Thus, there is a growing need for straightforward, accessible ways for the public to obtain relevant and important information or services based on climate predictions [2].

In recent years, large language models (LLMs) like ChatGPT-4 have provided a convenient means for the public to access specialized information [3]. With advanced natural language processing capabilities, LLMs can provide individuals with complex domain-specific knowledge. By processing simple language queries, LLMs allow users to obtain the necessary climate-related insights without requiring any training. The remarkable progress in knowledge integration and common-sense reasoning exhibited by LLMs has led to their exploration in various data-intensive fields, including scientific computing and financial forecasting, making climate science a promising area for LLM applications [4].

In the climate domain, several studies have attempted to apply LLMs to professional communication and information dissemination [5], [6]. For instance, the ClimSight project uses future forecast data from the Climate Modeling Intercomparison Project (CMIP) to provide agricultural recommendations through local climate services [7]. ClimateGPT, on the other hand, trains LLMs on climate science literature to generate more specialized climate knowledge and descriptive content [8]. However, most existing studies on climate LLMs focus on "descriptive output", in particular, on how to make LLMs generate content that reflects climate science terminology accurately [9]. While these models excel at interpretative tasks, the essential component of LLM responses for future climate scenarios lies in their ability to accurately predict future climate trends. Without this capacity to predict future climate factors reliably, LLM outputs may fall short of meeting the practical demands of climate adaptation and planning.

To understand what the general public, who is not trained in climate modeling and analysis, would be able to get from LLMs on climate predictions, this study aims to explore and assess the performance of LLMs in climate prediction tasks, particularly focusing on their ability to capture trends when generating future climate data We selected ChatGPT-4o, a representative LLM, for analysis and designed a series of experiments to evaluate its performance in generating short-term (15-day scale) and long-term (12-month scale) rainfall forecasts. The contribution of our study lies in advancing beyond existing climate LLM studies that focus primarily on descriptive output. We explore the predictive capabilities of LLMs, specifically their potential to generate accurate future climate trends with and without the assistance of domain-specific knowledge.

The structure of this article is as follows: Section 2 introduces the data used and the model employed as the climate expert model (EM). Section 3 describes the experiments conducted. Section 4 presents the comparative results. The discussion and conclusions are provided in Section 5.

## II. DATA AND METHOD

In this study, we focused on rainfall prediction at two different time scales: short-term (15-day scale) and long-term (12-month scale). We considered 15 cities across the contiguous United States: 'Washington DC' 'Tucson, AZ' 'Salt Lake City, UT' 'Reno, NV' 'Phoenix, AZ' 'Pensacola, FL' 'New York, NY' 'Mobile, AL' 'Forks, WA' 'El Paso, TX' 'Dallas, TX'

Yang Wang, Hassan A. Karimi are with the Geoinformatics Laboratory, School of Computing and Information, University of Pittsburgh, 135 N Bellefield Ave, Pittsburgh, PA 15213, USA (e-mail: yaw70@pitt.edu; hkarimi@pitt.edu).

'Chicago, IL' 'Birmingham, AL' 'Baton Rouge, LA' and 'Atlanta, GA' These cities represent varying levels of precipitation, covering high, medium, and low rainfall areas. For each city, we collected daily data on maximum temperature, minimum temperature, and precipitation, which were used as inputs for future rainfall prediction. Figure 1 shows the locations of the cities and their annual average rainfall. The data were sourced from the following publicly available databases:

- https://www.ncei.noaa.gov/pub/data/ghcn/daily/
- https://kilthub.cmu.edu/articles/dataset/Compiled_daily_temperature_and_precipitation_data_for_the_U_S_cities/7890488 [10]
- https://www.weather.gov/wrh/climate?wfo=sew.

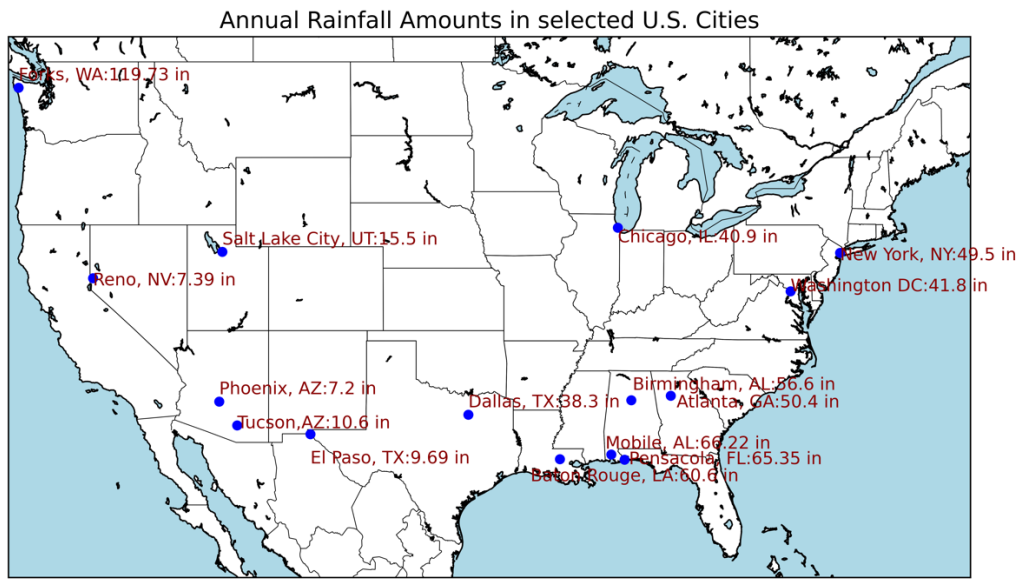

Figure 1 The locations of selected cities in the United States and their corresponding annual rainfall amounts.

For rainfall prediction, historical daily maximum temperature, minimum temperature, and precipitation were used as inputs. For temperature prediction, only minimum and maximum temperature were used as inputs. After preprocessing, the data from 1900 to 2022 were used, with 80% for training and 20% for validation.

We employed a two-layer LSTM model as our climate expert model (EM) for rainfall prediction. The model used an input window of 60-time steps, corresponding to 60 days for short-term prediction and 60 months for long-term prediction. The output window was set to 15-time steps for short-term prediction and 12-time steps for long-term prediction, respectively. The LSTM hidden size was set to 128, and a batch size of 64 was used. The model was trained over 500 epochs, with the Adam optimizer. The specific input features for rainfall prediction included historical minimum temperature, maximum temperature, and precipitation over the input window, predicting future precipitation. For temperature prediction, only the corresponding minimum and maximum temperatures were used as inputs.

To evaluate the model's predictive accuracy, we used data starting from September 30, 2023, as the baseline. For short-term predictions, we used the 60 days prior to this date to forecast daily rainfall from October 1 to October 15, 2023. For long-term predictions, we used the 60 months prior to October 2023 to forecast monthly rainfall from October 2023 to September 2024. This setup allowed us to evaluate the EM model's performance on both short-term and long-term rainfall forecasting tasks.

III. EXPERIMENTAL SETUP

In this study, to systematically evaluate the performance of LLMs in climate forecasting tasks and to analyze their ability to generate future climate data trends and patterns under various input scenarios, we designed several experiments. These experiments are intended to assess the LLM's ability to generate rainfall predictions with and without the support of specialized knowledge. The different experimental conditions also range from direct to indirect provision of expert information, allowing for a comprehensive analysis of GPT's potential in integrating climate data and generating forecasts. The specific experimental setups are as follows:

***Experiment 1: GPT-only Prediction***

In the first experiment, GPT-4o was tasked with independently generating rainfall predictions without any additional expert information or data input. This condition mainly assesses GPT's ability to produce short-term and long-term rainfall predictions based solely on its pre-trained knowledge. By analyzing its predictions, we aim to uncover GPT's inclination in judging future climate trends without external guidance. This experiment helps us understand the GPT's interpretation of climate events based on its common-sense knowledge. We used the following prompt sample:

| **Prompt Sample 1:** |
|---|
| *You are a climate data prediction system focused primarily on forecasting rainfall for selected cities. Your timestamp is September 30, 2023, meaning you only consider information available prior to this date. Please make a final forecast* |

> *based on your knowledge, including historical trends, regional variations, and potential future scenarios. For the time being, please ignore narrative responses; I am only interested in numerical results. Please predict for {city} during {October 1, 2023, to October 15, 2023}.*
>
> *—*
>
> *Please use the supplied data to predict the rainfall for the above period.*

### Experiment 2: GPT-EM Prediction

In the second experiment, we provided GPT-4o with rainfall predictions generated by a climate expert model (EM) and asked it to make further predictions based on this expert data. This condition is designed to examine whether GPT, after receiving direct forecast support from a specialized model, can effectively utilize this data to capture trends or adjust its predictions. By comparing GPT's predictions with and without the expert data support, we can analyze its ability to integrate external information. We used the following prompt sample:

> **Prompt Sample 2:**
>
> *You are a climate data prediction system focused primarily on forecasting rainfall for selected cities. Your timestamp is September 30, 2023, meaning you only consider information available prior to this date. I will provide you with a potential {daily} prediction for the period {October 1, 2023, to October 15, 2023} based on a deep learning model for the {city}. Please consider the results of the model and combine them with your knowledge to make a final forecast. For the time being, please ignore narrative responses; I am only interested in numerical results.*
>
> *— potential forecast —*
>
> *Period: {October 1, 2023, to October 15, 2023}*
>
> *Rainfall: {v1,v2,v3,v4,v5,v6,v7,v8,v9,v10,v11,v12,v13,v14,v15}*
>
> *—*
>
> *Please use the supplied data to predict the rainfall for the above period.*

### Experiment 3: GPT-Regional Climate Prediction

In the third experiment, we indirectly provided GPT with predictions of regional climate factors related to rainfall (minimum and maximum temperatures) and tasked it with generating rainfall predictions based on these climate factors. In this setup, GPT did not receive direct rainfall predictions but instead relied on relevant climate variables as hints to infer trends. This experiment aims to evaluate whether GPT can use indirect climate information to generate reasonable rainfall predictions and assess its ability to understand the relationships between climate variables in the absence of explicit rainfall data. We used the following prompt sample:

> **Prompt Sample 3:**
>
> *You are a climate data prediction system focused primarily on forecasting rainfall for selected cities. Your timestamp is September 30, 2023, meaning you only consider information available prior to this date. I will provide you with a potential {daily} prediction for the period {October 1, 2023, to October 15, 2023} based on a deep learning model for the {city}. These predictions include {daily} maximum and minimum temperatures. Please consider the relationship between these climate data and potential rainfall. Integrate this information with your knowledge to make a final prediction. For the time being, please ignore narrative responses; I am only interested in numerical results.*
>
> *— potential forecast —*
>
> *Period: {October 1, 2023, to October 15, 2023}*
>
> *Tmin: {v1,v2,v3,v4,v5,v6,v7,v8,v9,v10,v11,v12,v13,v14,v15}*
>
> *Tmax: {v1,v2,v3,v4,v5,v6,v7,v8,v9,v10,v11,v12,v13,v14,v15}*
>
> *—*
>
> *Please use the supplied data to predict the rainfall for the above period.*

### Experiment 4: GPT-Teleconnection Prediction

In the fourth experiment, to further explore GPT's understanding of large-scale climate indices, we provided it with global teleconnection indices, such as Nino3.4, the Pacific Decadal Oscillation (PDO), and the North Atlantic Oscillation (NAO), and instructed it to generate a 12-month-scale rainfall forecast based on this predictive information. These teleconnection indices are crucial variables in climate forecasting, revealing the long-term influence of large-scale climate patterns on variables such as rainfall. It should be noted that in this experiment, we used the actual values of these indices from October 2023 to September 2024 as inputs to serve as a comparison. We used the following prompt sample:

> **Prompt Sample 4:**
>
> *You are a climate data prediction system focused primarily on forecasting rainfall for selected cities. Your timestamp is September 30, 2023, meaning you only consider information available prior to this date. I will provide you with the Nino3.4, Pacific Decadal Oscillation (PDO), and North Atlantic Oscillation (NAO) indices for the prediction period {October 1, 2023, to October 15, 2023}. Please integrate this information, consider their climate teleconnection relationship with potential regional rainfall, and combine it with your own knowledge to make a final prediction. For the time being, please ignore narrative responses; I am only interested in numerical results.*

> *— potential forecast—*
>
> *Period: {October 1, 2023, to October 15, 2023}*
>
> *Nino3.4: {v1,v2,v3,v4,v5,v6,v7,v8,v9,v10,v11,v12,v13,v14,v15}*
>
> *PDO: {v1,v2,v3,v4,v5,v6,v7,v8,v9,v10,v11,v12,v13,v14,v15}*
>
> *NAO: {v1,v2,v3,v4,v5,v6,v7,v8,v9,v10,v11,v12,v13,v14,v15}*
>
> *—*
>
> *Please use the supplied data to predict the rainfall for the above period.*

### Baseline Comparison: 30-year Historical Average

To assess the accuracy and trend consistency of GPT's predictions, we used the average daily/monthly rainfall over the past 30 years as a baseline, against which we compared GPT's predictions generated under different experimental conditions. This baseline allows us to evaluate whether GPT's predictions tend to revert to historical averages and observe the differences in GPT's forecasts across various experimental settings.

We used Root Mean Square Error (RMSE) to measure the error between different predictions and actual rainfall. Additionally, we evaluated the accuracy and trend-capturing ability of the predictions using Pearson's correlation coefficient and Nash-Sutcliffe's efficiency coefficient.

## IV. RESULT

Figure 2 and Figure 3 present comparisons across different scales and scenarios. In this study, the EM model utilizes an LSTM structure, leveraging historical temperature and rainfall data to predict future rainfall. This time series-based deep learning model effectively captures the temporal dependencies within climate data, making it particularly suitable for detecting long-term trends and periodic patterns in meteorological data. As shown, the EM model achieves the best results in both long-term and short-term predictions. We are interested in examining the differences in LLM inferences when professional knowledge is directly incorporated compared to when it is not. However, when comparing Exp1 and Exp2, we found that providing GPT with relevant domain-specific knowledge and asking it to integrate this with its own knowledge did not significantly improve the results over relying solely on GPT. In short-term predictions, the average RMSE across all cities for Exp1 and Exp2 was 0.23 and 0.20, respectively, both notably higher than the EM model's 0.06. For long-term predictions, adding domain-specific knowledge actually increased RMSE compared to Exp1 and reduced the correlation coefficient. Although the provided domain knowledge might offer better predictive insights, results from both short-term and long-term predictions indicate that GPT's internal knowledge still plays a dominant role in its inferences.

We are also interested in observing what results GPT can provide when we input factors related to rainfall indirectly, rather than inputting rainfall data directly. Specifically, for long-term predictions, we added teleconnection-related factors (Exp4). In short-term predictions, the indirect provision of regional meteorological factors had limited impact on results; Exp3 showed an RMSE similar to Exp1 and Exp2, with a slightly improved correlation coefficient. However, in long-term predictions, whether adding regional factors or global teleconnection factors, GPT's results declined compared to when this knowledge was not added.

Comparing the results of different experiments with the 30-year historical average, we observed that GPT-generated predictions closely align with the 30-year average in both short-term and long-term forecasts. This similarity suggests that, in the absence of strong trends or notable anomalies, GPT tends to generate conservative predictions that resemble long-term statistical averages. This tendency is even more pronounced in long-term forecasts: we calculated the correlation coefficients between each experiment's results and the 30-year average, finding values of 0.86 for Exp1, 0.82 for Exp2, 0.76 for Exp3, and 0.62 for Exp4, while the EM model showed a lower correlation of only 0.59. This may be due to GPT's limited understanding of physical climate processes, leading it to default to safer, historically consistent forecasts. Additionally, this tendency reflects GPT's low sensitivity to complex climate patterns; in scenarios with limited data or high uncertainty, it relies on historical averages for robustness. However, this approach limits GPT's ability to capture potential trend shifts and the likelihood of extreme climate events.

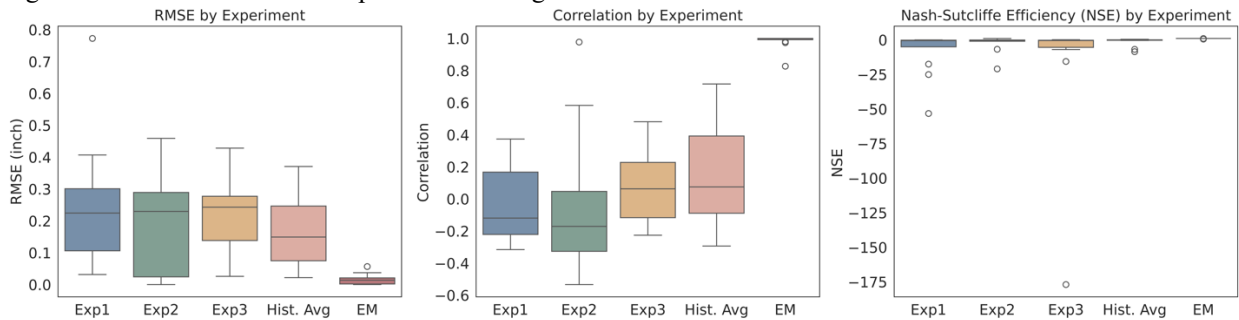

Figure 2 Performance comparison of rainfall prediction across different experimental conditions (short-term)

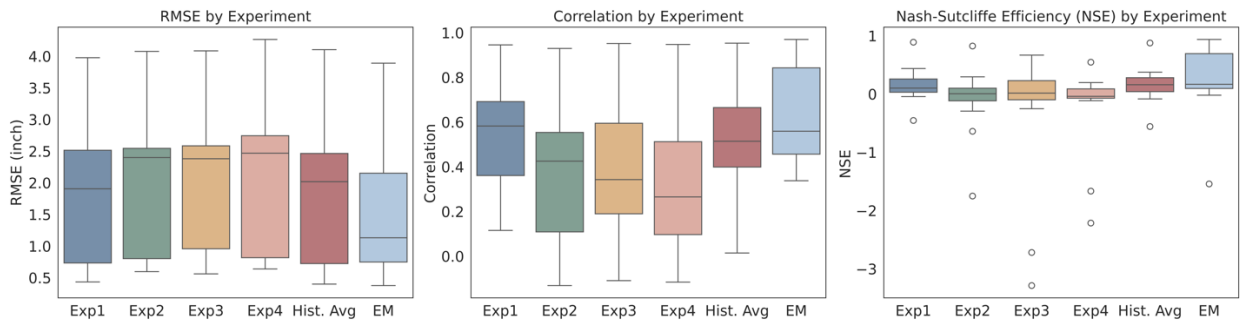

Figure 3 Performance comparison of rainfall prediction across different experimental conditions (long-term)

We further explored the differences between the results generated by GPT-4o and those from the EM model at each time point. In Figures 4 and 5, we compare the time series of short-term and long-term predictions. For the short-term predictions, we found that Exp2 tends to produce results closer to the multi-year average, especially during peak phases in the series. For peak values in the EM predictions, Exp2 noticeably dampened the magnitude of these peaks. For example, in Atlanta, there was an increase in rainfall on October 11 and 12, 2023, relative to the preceding days. The EM model effectively captured this upward trend, while Exp2 significantly reduced the peak values on these dates, bringing them down from the EM's 0.5-0.6 range to the 30-year average level of around 0.1-0.2. Similar patterns were observed on October 11 in Pensacola, October 6 and 14 in New York, and October 4 in Dallas. For peak values that were incorrectly predicted by the EM model, Exp2 also reduced their magnitude, as seen on October 15 in Phoenix and October 4 in Tucson.

We also found that Exp2's tendency to generate results close to the multi-year average was more pronounced in cities with higher rainfall. For instance, in Mobile and Baton Rouge, the results generated by Exp2 at each time point were closer to the 30-year average. A similar pattern can also be observed in the long-term predictions, where for all cities, Exp2's monthly-scale results were closer to the multi-year average at each time point. This tendency likely reflects GPT's preference for a more stable prediction aligned with the historical average, rather than adjusting fully to the extreme values suggested by the professional model. It also indicates that when encountering anomalies or extreme values, GPT is inclined to revert toward the average, leading to a smoothing effect in its predictions.

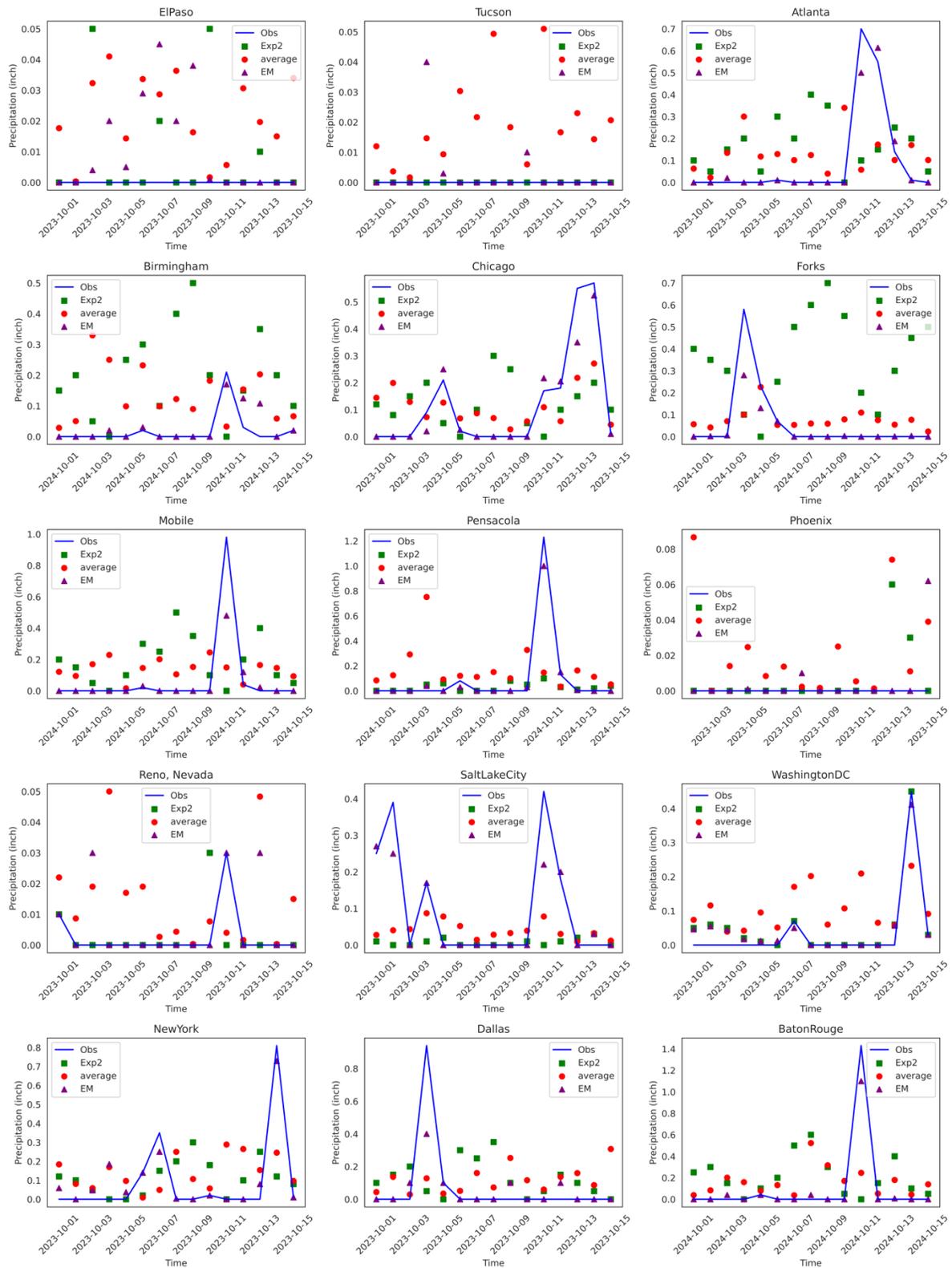

Figure 4 Time series comparison of short-term rainfall predictions across experimental conditions for different cities

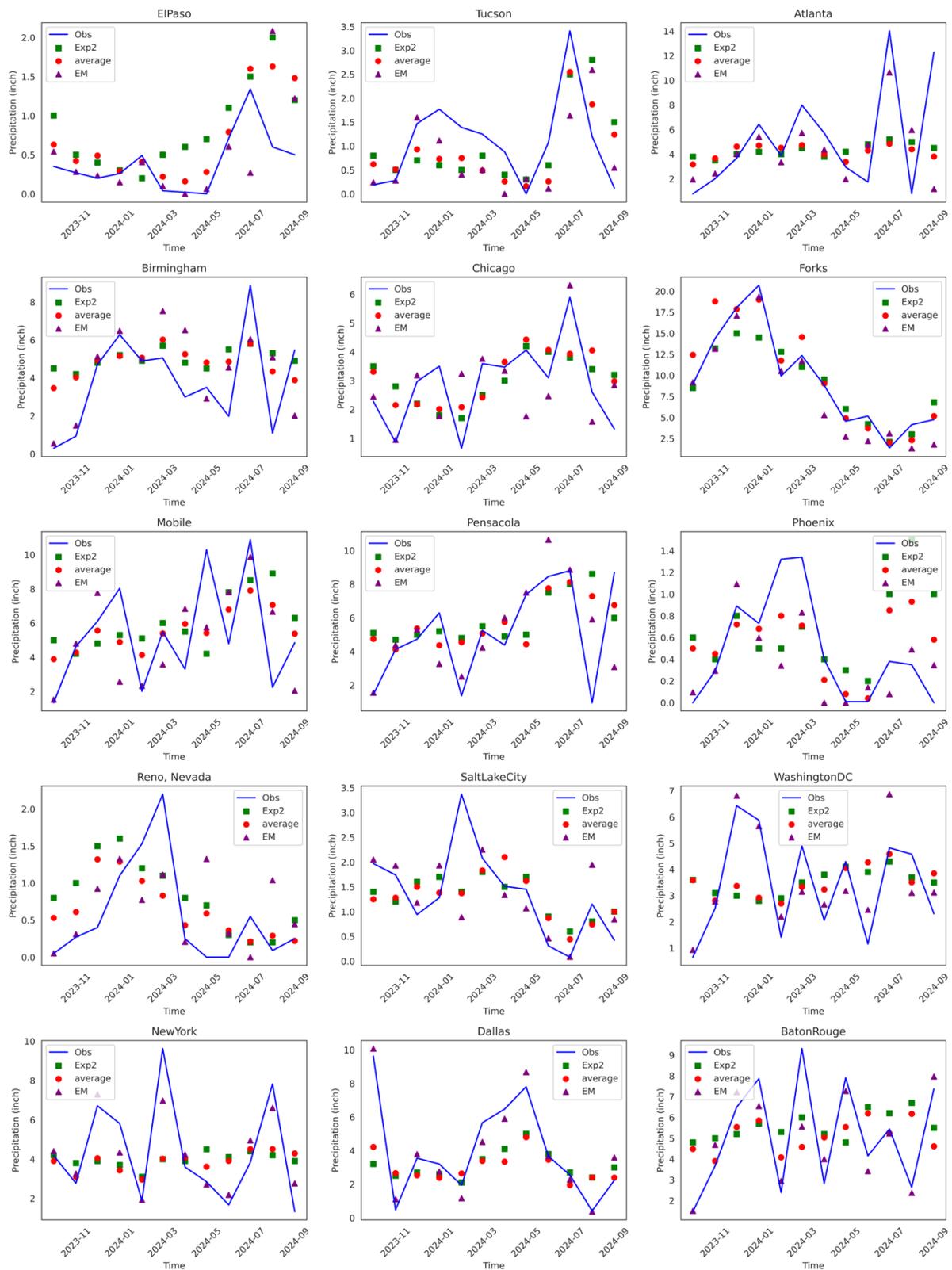

Figure 5 Time series comparison of long-term rainfall predictions across experimental conditions for different cities

**Fig. 4.** Comparison of predicted results and observation for different token setup and different lead time. A: whole test period; B: JFM prediction with lead time 6; C: AMJ prediction with lead time 14

## V. DISCUSSION AND FUTURE RESEARCH

As a large language model, GPT is fundamentally trained to learn language patterns from vast amounts of textual data, rather than physical processes. This means that it lacks the inherent physical constraints present in climate systems (such as energy conservation, atmospheric and ocean dynamics) and cannot consider causal relationships in predictions as physical models do. Thus, when predicting future rainfall trends, GPT primarily relies on "patterns in text" and "common-sense associations." When it fails to detect clear trend signals, it defaults to generating results aligned with historical averages— a relatively safe and conservative inference approach that avoids extreme or highly volatile predictions. This strategy reduces the risk of producing anomalous forecasts but also limits GPT's sensitivity to changing trends.

GPT's generation mechanism, based on language patterns, is more adept at producing general and trend-based content, yet often lacks sensitivity to rare or anomalous events. Extreme climate events are relatively infrequent in historical data, so without specialized training, GPT may lean toward predictions near the average, thus minimizing the risk of extreme error. This tendency causes GPT to smooth out results when dealing with extreme rainfall events, missing abnormal signals and downplaying the significance of extreme events. When directly provided with rainfall predictions from professional models (such as the EM model), GPT may not accurately interpret the data or effectively adjust its prediction strategy based on it. In other words, GPT is likely to treat this input as textual information, struggling to extract useful climate patterns or physical principles from it. This limitation diminishes its effectiveness in integrating specialized knowledge into its predictions.

In our study, we experimented with an alternative approach for long-term monthly-scale predictions. We calculated the standard deviation of historical rainfall for each calendar month. A high standard deviation indicates greater rainfall variability for that month, which corresponds to a higher prediction difficulty, while a low standard deviation suggests that rainfall levels are more consistent, indicating potentially lower prediction difficulty [11]. We used this standard deviation as a representation of potential uncertainty and input it into GPT along with the EM's predicted rainfall using the following prompt:

---

**Prompt Sample 5:**

*You are a climate data prediction system focused primarily on forecasting rainfall for selected cities. Your timestamp is September 30, 2023, meaning you only consider information available prior to this date. I will provide you with a potential {daily} prediction for the period {October 1, 2023 to October 15, 2023} based on a deep learning model for the {city}. The standard deviation here can be used as a measure of uncertainty. A smaller standard deviation indicates higher predictability, suggesting that my model's result has lower uncertainty. Conversely, a larger standard deviation indicates greater difficulty in prediction, meaning higher uncertainty in my model's results. Please focus on this measure of uncertainty, and combine it with your knowledge, such as historical trends, to make the final prediction. Please consider the results of the model and combine them with your knowledge to make a final forecast. For the time being, please ignore narrative responses; I am only interested in numerical results.*

*— potential forecast—*

*Period: {October 1, 2023 to October 15, 2023}*

*Rainfall:*
*{v1,v2,v3,v4,v5,v6,v7,v8,v9,v10,v11,v12,v13,v14,v15}*

*Standard Deviation:*
*{s1,s2,s3,s4,s5,s6,s7,s8,s9,s10,s11,s12,s13,s14,s15}*

*—*

*Please use the supplied data to predict the rainfall for the above period.*

---

This setup provided improved results. As shown in Figure 6, after adding the STD information, GPT's results are now close to those of the EM, with a significant improvement compared to Exp2. We present the time series comparison for the three cities with the most significant RMSE improvement in Figure 7. We observed that after adding this uncertainty information, GPT tended to provide results closer to those of the EM. However, we did not achieve the expected outcome where GPT would adjust the predictions with greater potential uncertainty.

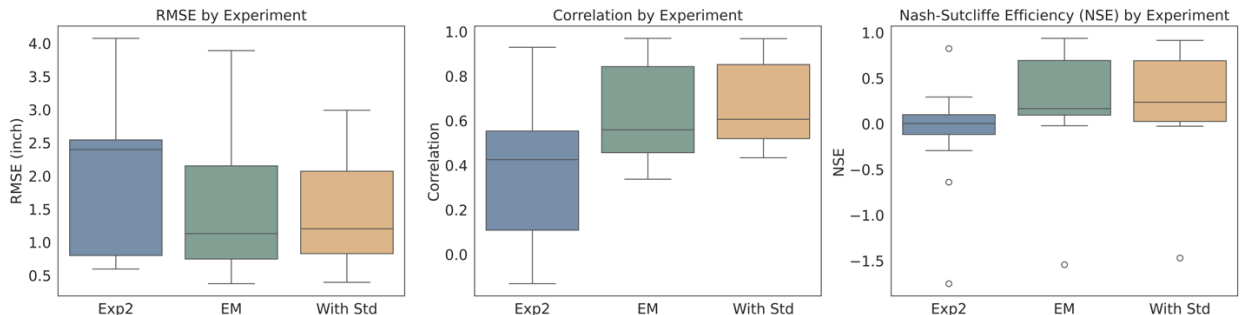

Figure 6 Performance comparison of rainfall prediction across Exp2, EM and another experiment of adding standard deviation

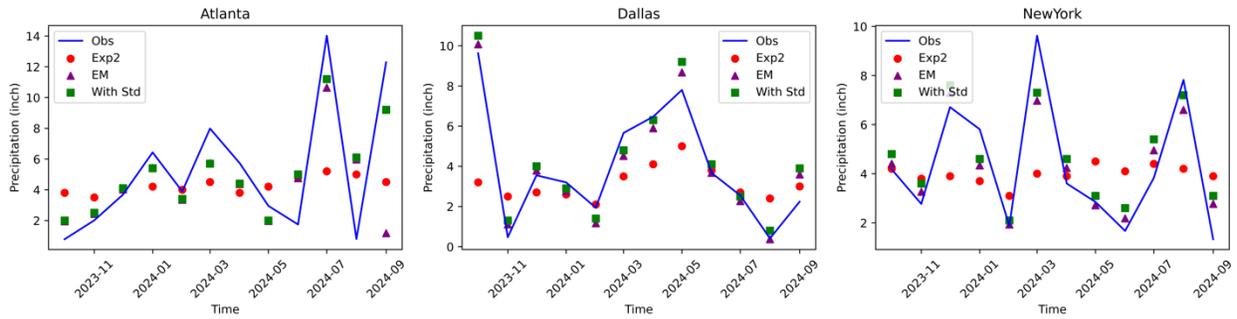

Figure 7 Time Series Comparison of long-term Rainfall Predictions across Exp2, EM and another experiment of adding standard deviation

A potential framework to integrate GPT's knowledge with EM's predictive insights is to leverage their strengths across different uncertainty intervals, resulting in more reliable predictions. By having the EM model provide uncertainty information along with its predictions, we can effectively combine GPT's conservative prediction tendencies with the high-precision predictions of the physical model. Specifically, the EM model's predictions can use uncertainty information to define "confidence regions." In low-uncertainty regions, GPT can directly adopt the EM model's predictions, as the EM model is generally reliable in these areas. In high-uncertainty regions, however, GPT can use its own knowledge to generate relatively conservative predictions, defaulting to historical averages or stable values to mitigate reliance on extreme fluctuations. By positioning the EM model as the "high-confidence predictor" and GPT as the "smoothing factor" in high-uncertainty scenarios, this strategy balances model accuracy with stability, enhancing overall prediction reliability. In this study, we analyzed only a limited set of cities and time series, which imposes certain constraints. Additionally, we provided domain knowledge through direct prompts. In the future, optimization strategies such as knowledge distillation, prompt engineering, or multi-task learning could be explored to improve GPT's understanding and handling of climate data [12]. For instance, incorporating a multi-task learning framework might allow GPT to autonomously learn the weights and importance of climate factors when processing climate data, potentially helping it to better integrate domain-specific knowledge and demonstrate greater potential in climate forecasting.

## VI. Conclusion

This study investigates the ability of large language models (LLMs), specifically ChatGPT-4o, to provide future climate information, focusing on rainfall prediction accuracy. We utilized a 2-layer LSTM model as an Expert Model (EM), assuming it could generate reliable future predictions. Through a series of experiments, we compared ChatGPT-4's rainfall predictions under varying conditions: relying solely on its internal knowledge, directly receiving rainfall predictions from the EM, and indirectly inferring rainfall from other related factors predicted by the EM. These experiments evaluated both short-term (15-day, daily scale) and long-term (12-month, monthly scale) prediction capabilities. The results reveal that ChatGPT-4 consistently prioritizes stable predictions closely aligned with historical averages, regardless of whether it integrates additional information from the EM. This tendency highlights the LLM's inherent bias towards conservative outputs, which may limit its effectiveness in capturing dynamic or extreme variations in climate scenarios.

**Reference:**
[1] S. D. Campbell and F. X. Diebold, "Weather forecasting for weather derivatives," *J Am Stat Assoc*, vol. 100, no. 469, pp. 6–16, Mar. 2005, doi: 10.1198/016214504000001051.
[2] C. Hewitt, S. Mason, and D. Walland, "The global framework for climate services," Dec. 2012. doi: 10.1038/nclimate1745.
[3] OpenAI, "GPT-4 Technical Report."
[4] M. Leippold, "Thus spoke GPT-3: Interviewing a large-language model on climate finance," *Financ Res Lett*, vol. 53, May 2023, doi: 10.1016/j.frl.2022.103617.
[5] H. Zhu and P. Tiwari, "Climate Change from Large Language Models."
[6] M. Kraus *et al.*, "Enhancing Large Language Models with Climate Resources." [Online]. Available: www.climatewatchdata.org/
[7] N. Koldunov and T. Jung, "Local climate services for all, courtesy of large language models," Dec. 01, 2024, *Nature Publishing Group*. doi: 10.1038/s43247-023-01199-1.
[8] D. Thulke *et al.*, "ClimateGPT: Towards AI Synthesizing Interdisciplinary Research on Climate Change," 2024. [Online]. Available: https://huggingface.co/eci-io/
[9] N. Webersinke, M. Kraus, J. A. Bingler, and M. Leippold, "CLIMATEBERT: A Pretrained Language Model for Climate-Related Text," 2022. [Online]. Available: www.github.com/climatebert/language-model
[10] Y. Lai and D. A. Dzombak, "Use of Historical Data to Assess Regional Climate Change", doi: 10.1175/JCLI-D-18.

[11] M. Zhang, J. D. Rojo-Hernández, L. Yan, Ó. J. Mesa, and U. Lall, "Hidden Tropical Pacific Sea Surface Temperature States Reveal Global Predictability for Monthly Precipitation for Sub-Season to Annual Scales," *Geophys Res Lett*, vol. 49, no. 20, pp. 1–9, 2022, doi: 10.1029/2022GL099572.

[12] Q. Wu *et al.*, "AutoGen: Enabling Next-Gen LLM Applications via Multi-Agent Conversation," 2023.

.